\documentclass[11pt,letterpaper]{article}
\usepackage[letterpaper]{geometry}
\usepackage{acl2012}
\usepackage{times}
\usepackage{latexsym}
\usepackage{amsmath}
\usepackage{multirow}
\usepackage{url}

\usepackage{graphicx}
\usepackage{listings}
\usepackage{subcaption}
\usepackage{float}
\usepackage{color}

\makeatletter
\newcommand{\@BIBLABEL}{\@emptybiblabel}
\newcommand{\@emptybiblabel}[1]{}
\makeatother

\newcommand{\tabincell}[2]{\begin{tabular}{@{}#1@{}}#2\end{tabular}}

\title{Bidirectional Tree-Structured LSTM with Head Lexicalization}

\author{Zhiyang Teng \and
  Yue Zhang \\
   Singapore University of Technology and Design\\
  {\tt zhiyang\_teng@mymail.sutd.edu.sg} \\
  {\tt yue\_zhang@sutd.edu.sg} \\}

\date{}

\begin{document}
\maketitle
\begin{abstract}
 %Long short-term memory (LSTM) has been shown highly effective for modeling non-local features in sentences. 
%Recently, 
Sequential LSTM has been extended to model tree structures, giving competitive results for a number of tasks. 
Existing methods model constituent trees  by bottom-up combinations of constituent nodes, making direct use of input word information only for leaf nodes. 
This is different from sequential LSTMs, which contain reference to input words for each node. 
In this paper, we propose a method for automatic head-lexicalization for tree-structure LSTMs, propagating head words from leaf nodes to every constituent node. 
In addition, enabled by head lexicalization, we %build 
%we follow bidirectional sequential LSTM and 
build a tree LSTM in the top-down direction, which corresponds to bidirectional sequential LSTM structurally. Experiments show that both extensions give better representations of tree structures. Our final model gives the best results on the Standford Sentiment Treebank and highly competitive results on the TREC question type classification task.
\end{abstract}

\section{Introduction}

%Sequences and trees are the two most common structures in natural langauge processing (NLP).  
%For example,  language modeling \cite{chen1996empirical,koehn2009statistical},
%Part-of-Speech (PoS) tagging \cite{postagging,lafferty2001conditional} and named entity recognition \cite{ner} have been tackled as sequence labeling problems, 
%while syntactic parsing \cite{collins2003head} and structured machine translation \cite{chiang:2005:ACL} have been solved as tree structure problems.

Both sequence structured and tree structured neural models have been applied to NLP problems.  
Seminal work employs convolutional neural network \cite{collobert2008unified}, recurrent neural network  \cite{elman1990finding,mikolov2010recurrent} and recursive neural network  \cite{SocherEtAl2011:RNN} for sequence and tree modeling. 
Recently, Long Short-Term Memories (LSTM) have received increasing research attention, giving significantly improved accuracies in a variety of sequence tasks 
\cite{sutskever2014sequence,bahdanau2014neural} compared to vanilla recurrent neural networks. 
Addressing diminishing gradients effectively, they have been extended to tree structures, achieving promising 
results for tasks such as syntactic language modeling \cite{topdown}, sentiment analysis \cite{li2015tree,zhu15,le15,tai15} and relation extraction \cite{bidir}. 

%\begin{figure}[!t]
%    \begin{center}
%        \includegraphics[scale=0.3]{con+dep.jpg}
%        \caption{{\bf Tree Strcutures In NLP.}}
%        \label{fig:condep}
%    \end{center}
%\end{figure}

According to the node type, typical tree structures in NLP can be categorized to \emph{constituent} trees and \emph{dependency} trees.
%, which are illustrated in Figure \ref{fig:condep}(a) and \ref{fig:condep}(b), respectively. 
%The figure shows the syntactic tree structures of a sentence ``Jonh visited Mary this afternoon''. Parts-of-speech such as the proper noun (NNP) and the past tense verb (VBD) are marked above each word. 
%As shown in the figure, 
A salient difference between the two types of tree structures is in the node. While dependency tree nodes are \emph{input words} themselves, constituent tree nodes represent \emph{syntactic constituents}. 
Only leaf nodes in constituent trees correspond to words. Though LSTM structures have been developed for both types of trees above, we investigate constituent trees in this paper. 
There are three existing methods for constituent tree LSTM \cite{zhu15,tai15,le15}, which make essentially the same extension from sequence structure LSTMs.
We take the method  of \newcite{zhu15} as our baseline. 

\begin{figure}[!t]
    \begin{center}
        \includegraphics[width=0.5\textwidth]{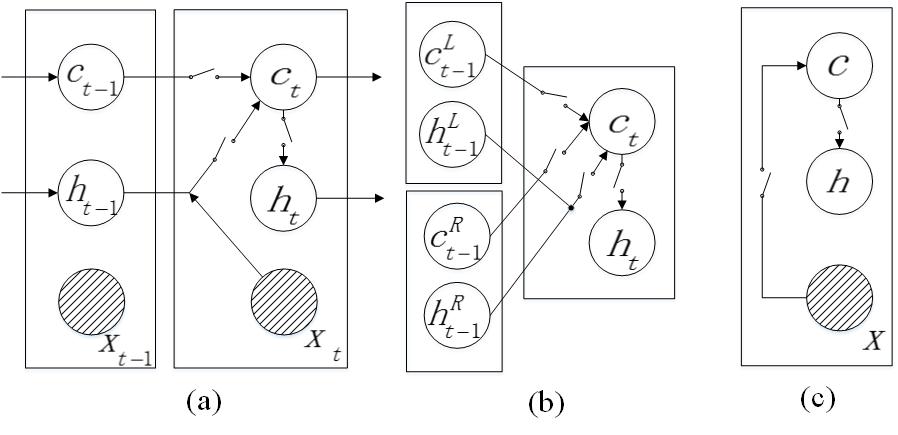}
        \caption{{Topology of sequential and tree LSTMs. (a) nodes in sequential LSTM; (b) non-leaf nodes in tree LSTM; (c) leaf nodes in tree LSTM.  Shaded nodes represent lexical input vectors. White nodes represent hidden state vectors.}}
        \label{fig:treelstm}
    \end{center}
    \vspace{-2em}
\end{figure}

A contrast between the sequence structured LSTM of \newcite{hochreiter1997long} and the tree-structured  LSTM of \newcite{zhu15} is shown in Figure \ref{fig:treelstm}, which illustrates the input ($x$), cell ($c$) and hidden ($h$) nodes at a certain time step $t$. 
The most important difference between Figure \ref{fig:treelstm}(a) and Figure \ref{fig:treelstm}(b) is the branching factor. 
While a cell in the sequence structure LSTM depends on the single previous hidden node, a cell in the tree-structured LSTM depends on a left hidden node and a right hidden node. 
Such tree-structured extension of the sequence structure LSTM assumes that the constituent tree is binarized, building hidden nodes from the input words in the bottom-up direction. 
The leaf node structure in shown in Figure \ref{fig:treelstm}(c). 

A second salient difference between the two types of LSTMs is the modeling of input words. 
While each cell in the sequence structure LSTM directly depends on its corresponding input word, only leaf cells in the tree 
structure LSTM directly depend on  corresponding input words. This corresponds well to the constituent tree structure, where there is no direct association between non-leaf constituent nodes and input words. 
However, it leaves the tree structure a degraded version of a perfect binary-branching variation of the sequence-structure LSTM, with one important source of information (i.e. words)  missing in forming a cell. 

\begin{figure}[!t]
    \begin{center}
        \includegraphics[scale=0.3]{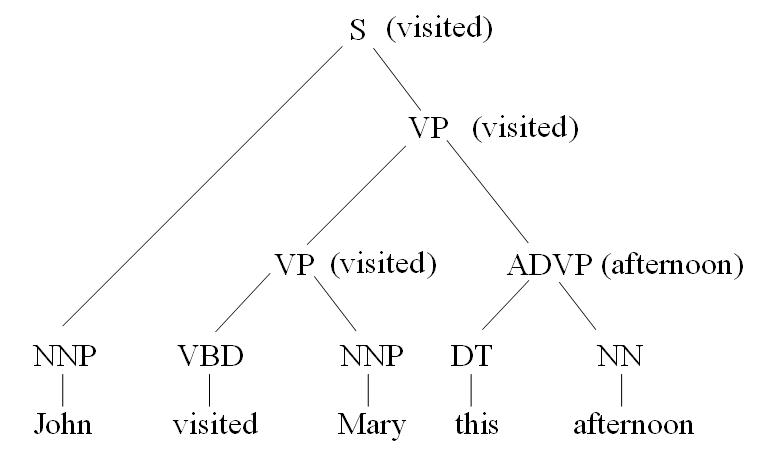}
        \caption{{Head-Lexicalized Constituent Tree.}}
        \label{fig:headour}
    \end{center}
    \vspace{-2em}
\end{figure}

We fill this gap by proposing an extension to the tree LSTM model, injecting lexical information into every node in the tree.
Our method takes inspiration from work on head-lexicalization, which shows that each node in a constituent tree structure is governed by a head word. 
As shown in Figure \ref{fig:headour}, the head word for the verb phrase ``visited Mary'' is ``visited'', and the head word of the adverb phrase ``this afternoon'' is ``afternoon''. 
Research has shown that head word information can significantly improve the performance of syntactic parsing \cite{collins2003head,clark-curran:2004:ACL}. 
Correspondingly, we use head lexical information of each constituent word as the input node $x$ for calculating the corresponding cell $c$ in Figure \ref{fig:treelstm}(b). 

Traditional head-lexicalization relies on specific rules \cite{collins2003head}, typically extracting heads from constituent treebanks according to certain grammar formalisms. 
For better generalization,  we use a neural attention mechanism to derive head lexical information automatically, 
rather than relying on linguistic head rules to find the head lexicon of each constituent, which is language- and formalism-dependent. 

Based on such head lexicalization, we make a bidirectional extension of the tree structured LSTM, propagating information in the top-down direction as well as the bottom-up direction. 
This is analogous to the bidirectional extension of sequence structure LSTMs, which are commonly used for NLP tasks such as speech recognition \cite{graves2013hybrid}, 
sentiment analysis \cite{tai15,li2015tree} and machine translation \cite{sutskever2014sequence,bahdanau2014neural}
tasks. 

Results on a standard sentiment classification benchmark and a question type classification benchmark show that our tree LSTM structure gives significantly better accuracies compared with the method of \newcite{zhu15}. We achieve the best reported results for sentiment classification. 
Interestingly, the head lexical information that is learned automatically from the sentiment treebank consists of both syntactic head information 
and key sentiment word information. 
This shows the advantage of automatic head-finding as compared with rule-based head lexicalization. 
We make our code available under GPL at https://github.com/XXX/XXX.

\section{Related Work}
LSTM \cite{hochreiter1997long} is a variation of RNN \cite{elman1990finding} to solve vanishing gradients in training. 
It has been widely adopted for NLP tasks, such as parsing \cite{dyer-EtAl:2015:ACL-IJCNLP} and machine translation \cite{bahdanau2014neural}. We take the standard LSTM with peephole connections \cite{gers2000recurrent} as our baseline, which models sequences. 

There has been extensions of sequence-structured LSTMs for modeling trees \cite{tai15,le15,zhu15}. The idea is similar to extending recursive neural network to recurrent neural network, where gates are used in recursive structures for alleviating diminishing gradients. While \newcite{tai15} investigated tree-structured LSTMs for both dependency structures and constituent structures, and both for unrestricted trees and M-nary trees, \newcite{zhu15} and \newcite{le15} focused on binary constituent trees. As discussed earlier, none of these existing methods make direct use of lexical input for composing constituent vectors. We take \newcite{zhu15} as our baseline. 

Bidirectional information has been leveraged to extend sequence LSTM \cite{graves2013hybrid}. On the other hand, the aforementioned tree-LSTM models work bottom-up, without information flew from parents to children. \newcite{topdown} built a top-down tree-LSTM  for dependency trees, but without a bottom-up component. \newcite{paulus2014global} made use of bidirectional information on binary trees, but for recursive neural networks only. The closest in spirit to our method, \newcite{bidir} adopted a bidirectional Tree LSTM model to jointly extract named entities and relations under \emph{dependency} tree structure. 
For \emph{constituent} tree structures, however, their model does not work due to lack of word inputs on non-leaf constituent nodes, and in particular  the root node. 
Our head lexicalization allows us to investigate the top-down constituent Tree LSTM. To our knowledge, we are the first to report a bidirectional constituent Tree LSTM.

\section{Baselines}

A sequence-structure LSTM  estimates a sequence of hidden \textbf{state vectors} given a sequence of \textbf{input vectors}, 
through the calculation of a sequence of hidden \textbf{cell vectors} using a gate mechanism. For NLP, the input vectors are typically word embeddings \cite{mikolov2013efficient}, but can also include PoS embeddings, character embeddings or other types of information. 
For notational convenience, we refer to the input vectors as \textbf{lexical vectors}. 

Formally, given an input vector sequence $x_1, x_2, \dots, x_n$, each state vector $h_t$ is estimated from the Hadamard product of a cell vector $c_t$ and a corresponding \textbf{output gate vector} $o_t$ 
\begin{equation}
\label{eq:ht}
h_t = o_t \otimes \tanh(c_t)
\end{equation}
Here the cell vector depends on both the previous cell vector $c_t$, and a combination of the previous state vector $h_{t-1}$, the current input vector $x_t$:

\begin{equation}
  c_t = f_t \otimes c_{t-1} + i_t  \otimes g_t 
\end{equation}
\begin{equation}
  g_t = \tanh ( W_{xg}  x_{t} + W_{hg} h_{t-1}  + b_g ) 
\end{equation}

The combination of $c_{t-1}$ and $g_t$ are controlled by the Hadamard product between a \textbf{forget gate} vector $f_t$ and a \textbf{input gate} vector $i_t$, respectively. 
The gates $o_t$, $f_t$ and $i_t$ are defined as follows
\begin{equation}
    \begin{split}
        i_t &= \sigma ( W_{xi} x_{t}  + W_{hi}  h_{t-1} + W_{ci}  c_{t-1}   + b_i ) \\
        f_t &= \sigma ( W_{xf} x_{t}  + W_{hf}  h_{t-1} + W_{cf}  c_{t-1}   + b_f ) \\
        o_t &= \sigma (W_{xo} x_{t}  + W_{ho}  h_{t-1} + W_{co}  c_{t}   + b_o),  \\
    \end{split}
\end{equation}
where $\sigma$ is the sigmoid function.  $W_{xg}$, $W_{hg}$, $b_g$, $W_{xi}$, $W_{hi}$, $W_{ci}$, $b_i$,  $W_{xf}$, $W_{hf}$, $W_{cf}$, $b_f$, $W_{xo}$, $W_{ho}$, $W_{co}$ and $b_o$
are model parameters.

The bottom-up Tree LSTM of \newcite{zhu15} extends the left-to-right sequence LSTM by splitting the \emph{previous} state vector $h_{t-1}$ into a \emph{left child} state vector $h_{t-1}^L$ and a \emph{right child} state vector $h_{t-1}^R$, and the previous cell vector $c_{t-1}$ into a left child cell vector $c_{t-1}^L$ and a right child cell vector $c_{t-1}^R$, calculating $c_t$ as 
\begin{equation}
\label{eq:ct}
  c_t  = f_t^L \otimes  c_{t-1} ^L  + f_t^R \otimes  c_{t-1} ^R + i_t \otimes  g_t, 
\end{equation}
and the input/output gates $i_t$/$o_t$ as 
\begin{equation}
\begin{split}
 i_t  &= \sigma \Big( \sum_{N \in \{L, R\}} ( W_{hi} ^ {N} h_{t-1} ^N + W_{ci} ^ {N} c_{t-1} ^N) + b_i \Big) \\
 o_t &= \sigma \Big(\sum_{N \in \{L, R\}} W_{ho} ^ {N} h_{t-1} ^N + W_{co}  c_{t}  + b_o\Big) \\
\end{split}
\end{equation}
 The forget gate $f_t$ is split into $f_t^L$ and $f_t^R$ for regulating $c_{t-1}^L$ and $c_{t-1}^R$, respectively: 
\begin{equation}
 \begin{split}
 f_t^L &= \sigma \Big( \sum_{N \in \{L, R\}} ( W_{hf_l} ^ {N} h_{t-1} ^N + W_{cf_l} ^ {N} c_{t-1} ^N) + b_{f_l} \Big) \\
 f_t^R &= \sigma \Big(  \sum_{N \in \{L, R\}} ( W_{hf_r} ^ {N} h_{t-1} ^N + W_{cf_r} ^ {N} c_{t-1} ^N) + b_{f_r} \Big) \\
\end{split}
\end{equation}
$g_t$ depends on both $h_{t-1}^L $ and $h_{t-1}^R$, but as shown in  Figure \ref{fig:treelstm} (b), it does not depend on $x_t$
\begin{equation}
 g_t = \tanh \Big(  \sum_{N \in \{L, R\}} W_{hg} ^ {N} h_{t-1} ^N  + b_g \Big) 
\end{equation}

Finally, the hidden state vector $h_t$ is calculated in the same way as in the sequential LSTM model shown in Equation \ref{eq:ht}. 
$W_{hi} ^ {L}$, $W_{hi} ^ {R}$,   $W_{ci} ^ {L}$,  $W_{ci} ^ {R}$, $b_i$, $W_{ho} ^ {L}$, $W_{ho} ^ {R}$, $W_{co}$, $b_o$, 
$W_{hf_l} ^ {L}$, $W_{hf_l} ^ {R}$,  $W_{cf_l} ^ {L}$, $W_{cf_l} ^ {R}$,  $b_{f_l}$,   $W_{hf_r} ^ {L}$, $W_{hf_r} ^ {R}$,
$W_{cf_r} ^ {L}$, $W_{cf_r} ^ {R}$, $b_{f_r}$,  $W_{hg} ^ {L}$,  $W_{hg} ^ {R}$ and $b_g$ are model parameters.

%where $\sigma$ is the sigmoid function; $\otimes$ is the element-wise multiplication; 

\section{Our Model}
\subsection{Head Lexicalization}
\label{sec:headlex}
\begin{figure}[!t]
    \centering
    \begin{subfigure}[b]{0.19\textwidth}
         \includegraphics[width=\textwidth]{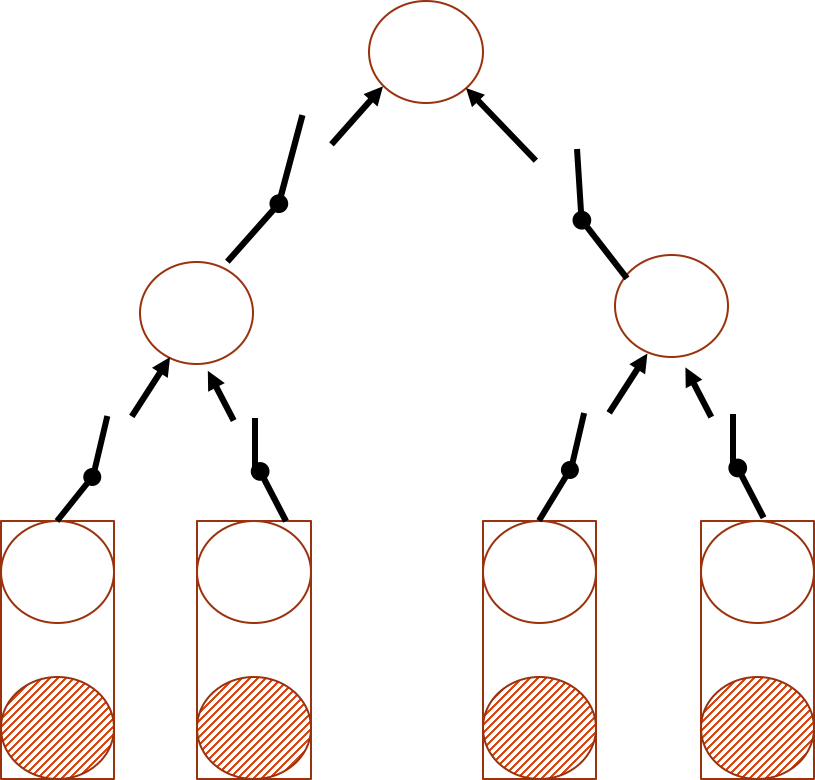}
         \caption{}
    \end{subfigure}
    \hspace{10pt}
    \begin{subfigure}[b]{0.21\textwidth}
        \includegraphics[width=\textwidth]{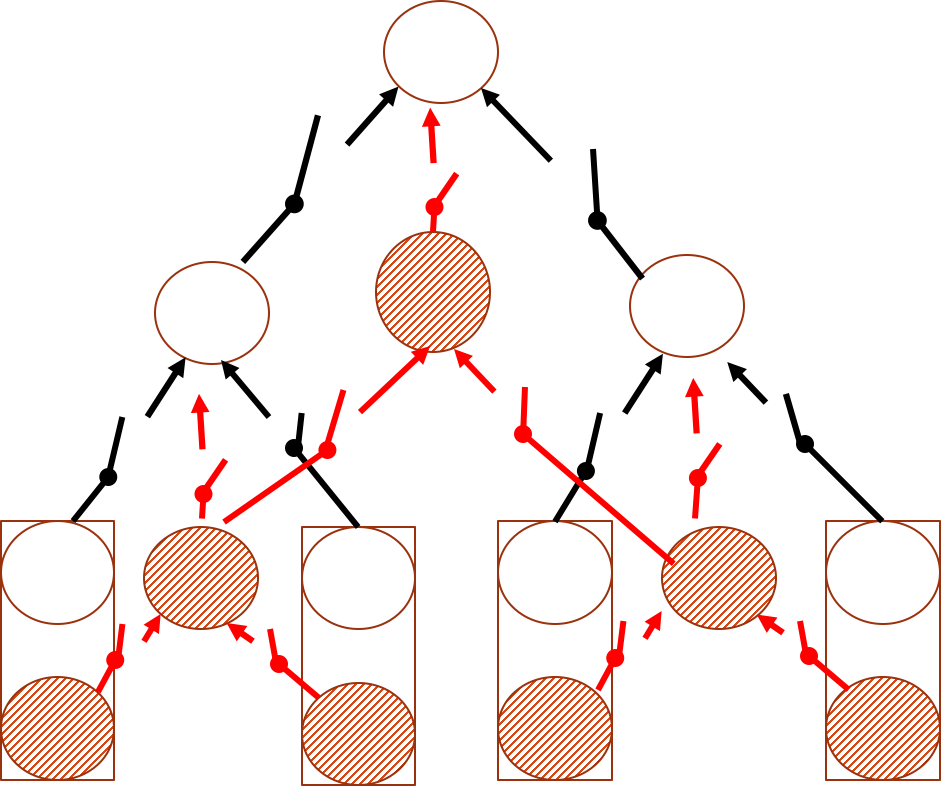}
        \caption{}
    \end{subfigure}
    \caption{{Contrast between \newcite{zhu15} (a)  and this paper (b). Shaded nodes represent lexical input vectors. White nodes represent hidden state vectors.}}
    \label{fig:headlex}
    \vspace{-0.5em}
\end{figure}

We introduce an input lexical vector $x_t$ to the calculation of each cell vector $c_t$ via a bottom-up head propagation mechanism. 
As shown in the shaded nodes in Figure \ref{fig:headlex} (b),  the head propagation mechanism stands relatively in parallel to  the cell propagation mechanism.  
In contrast, the method of \newcite{zhu15} in Figure \ref{fig:headlex} (a) does not have the input vector $x_t$ for non-leaf constituents.  
%\begin{figure}[!t]
%    \begin{center}
%        \includegraphics[width=0.5\textwidth]{headlex.jpg}
%        \caption{{Contrast between \newcite{zhu15} (a)  and this paper (b).}}
%        \label{fig:headlex}
%    \end{center}
%\end{figure}

There are multiple ways to choose a head lexicon for a given binary-branching constituent.  
One simple baseline is to choose the head lexicon of the left child as the head (left-headedness).
Correspondingly, an alternative is to use the right child for head lexicon. 
However, these simple baselines can bring less benefits  compared to linguistically motivated head finding, due to relatively less consistency in the governing head lexicons across variations of 
the same type of constituents with slightly different typologies. 

Rather than selecting head lexicons using manually-defined head-finding rules, which are language- and formalism-dependent \cite{collins2003head}, we cast head finding as 
a part of the neural network model, learning the head lexicon of each constituent by a gated combination of head lexicons of its two children\footnote{In this paper, we work on binary trees only, which is a common form for CKY and shift-reduce parsing. Typicial binarization methods, such as head binarization \cite{klein2003accurate}  , also rely on specific head-finding rules.}.  Formally, 
\begin{equation}
 x_t = z_t \otimes x_{t-1}^L  + ( 1 - z_t ) \otimes x_{t-1}^R, 
\end{equation}
where $x_t$ represents the head lexicon vector of the current constituent, $x_{t-1}^L$ represents the head lexicon of its left child constituent, and $x_{t-1}^R$ represents the head lexicon of
its right child constituent. The gate $z_t$ is calculated based on $x_{t-1}^L$ and $x_{t-1}^R$, 
\begin{equation}
 z_t = \sigma ( W_{zx}^L x_{t-1}^L  + W_{zx}^R x_{t-1} ^R  + b_z)
\end{equation}
Here $W_{zx}^L$, $W_{zx}^R$ and $b_z$ are model parameters. 

\subsection{Lexicalized Tree LSTM}
\label{subsec:lextree}
Given head lexicon vectors for nodes, the Tree LSTM of \newcite{zhu15} can be extended by leveraging $x_t$ in calculating the corresponding $c_t$. 
In particular, $x_t$ is used to estimate the input ($i_t$), output ($o_t$) and forget ($f_t^R$ and $f_t^L$) gates: 
\begin{equation}
    \begin{split}
        i_t &= \sigma \Big(   \mathbf{W_{xi} x_{t}} +\\&\sum_{N \in \{L, R\}} ( W_{hi} ^ {N} h_{t-1} ^N + W_{ci} ^ {N} c_{t-1} ^N) + b_i \Big) \\
        f_t^L &= \sigma \Big( \mathbf{W_{xf} x_{t}} +\\&\sum_{N \in \{L, R\}} ( W_{hf_l} ^ {N} h_{t-1} ^N + W_{cf_l} ^ {N} c_{t-1} ^N) + b_{f_l} \Big) \\
        f_t^R &= \sigma \Big( \mathbf{W_{xf} x_{t}} +\\& \sum_{N \in \{L, R\}} ( W_{hf_r} ^ {N} h_{t-1} ^N + W_{cf_r} ^ {N} c_{t-1} ^N) + b_{f_r} \Big) \\
        o_t &= \sigma \Big(\mathbf{W_{xo}  x_{t}} +\\&\sum_{N \in \{L, R\}} W_{ho} ^ {N} h_{t-1} ^N + W_{co}  c_{t}  + b_o\Big)  \\
    \end{split}
\end{equation}
In addition, $x_t$ is also used in computing $g_t$, 
\begin{equation}
  g_t = \tanh \Big( \mathbf{W_{xg}  x_{t}} +  \sum_{N \in \{L, R\}} W_{hg} ^ {N} h_{t-1} ^N + b_g \Big) 
\end{equation}
With the new definition of $i_t$, $f_t^R$, $f_t^L$ and $g_t$, the computing of $c_t$ remains the same as the baseline Tree LSTM model  as shown in Equation \ref{eq:ct}. 
Similarly, $h_t$ remains the Hadamard product of $c_t$ and the new $o_t$ as shown in Equation \ref{eq:ht}. 

In this model, $W_{xi}$, $W_{xf}$, $W_{xg}$ and $W_{xo}$ are newly-introduced model parameters. The use of $x_t$ in computing the gate and cell values are consistent with those 
in the baseline sequential LSTM. 
\subsection{Bidirectional Extensions}
Given a sequence of input vectors [$x_1$, $x_2$, $\dots$, $x_n$],  a bidirectional \emph{sequential} LSTM \cite{graves2013hybrid} computes two sets of hidden state vectors, $[\tilde{h}_1$, $\tilde{h}_2$, $\dots$, $\tilde{h}_n]$ and 
$[\tilde{h}_{n}^{\prime} $, $\tilde{h}_{n-1}^{\prime} $, $\dots$, $\tilde{h}_1^{\prime}]$ in the left-to-right and the right-to-left directions, respectively.  
The final hidden state $h_i$ of the input $x_i$
is the concatenation of the corresponding state vectors in the two LSTMs, 
\begin{equation}
  h_i =  \tilde{h}_i  \oplus \tilde{h}_{n-i+1}^{\prime} 
\end{equation}
The two LSTMs can share the same model parameters or use different parameters. We  choose the latter in our baseline experiments. 

We make a bidirectional extension to the Lexicalized tree LSTM in Section \ref{subsec:lextree} by following the sequential baseline above, adding an additional set of hidden state vectors in the top-down direction. 
Different from the bottom-up direction,  each hidden state in the top-down LSTM has exactly one predecessor. 
In fact, the path from the root of a tree down to any node forms a sequential LSTM. 

\begin{figure}[!t]
    \begin{center}
        \includegraphics[width=0.25\textwidth]{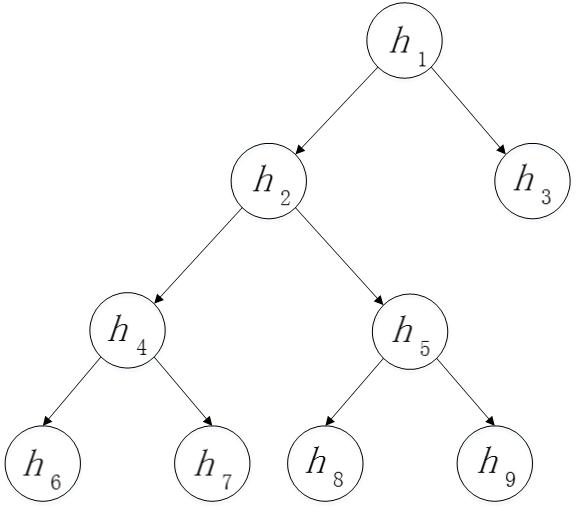}
        \caption{{ Top-down tree LSTM. }}
        \label{fig:topdown}
    \end{center}
    \vspace{-1em}
\end{figure}

Note, however, that two different sets of model parameters are  used when the current node is the left and the right child of its predecessor, respectively.
Denoting the two sets of parameters as $\mathbf{U}_L$ and $\mathbf{U}_R$, respectively,
the hidden state vector $h_7$ in Figure \ref{fig:topdown} is  calculated from the hidden state vector $h_1$ using the parameter set sequence $[\mathbf{U}_L, \mathbf{U}_L, \mathbf{U}_R]$. 
Similarly, $h_8$ is calculated from $h_1$ using  $[\mathbf{U}_L, \mathbf{U}_R, \mathbf{U}_L]$.  
At each step $t$, the computing of $h_t$ follows the sequential LSTM model:
\begin{equation}
    \begin{split}
        h_{t} &=  o_t \otimes \tanh (c_{t-1}) \\
        c_{t}  &= f_{t} \otimes  c_{t-1}   + i_{t} \otimes  g_{t} \\
        g_{t} &= \tanh ( W_{xg\downarrow}^N  x_{t-1} + W_{hg\downarrow}^N  h_{t-1}  + b_{g\downarrow} ^ N ) \\
    \end{split}
\end{equation}
With the gate values being defined as: 
\begin{equation}
    \begin{split}
        i_{t} &= \sigma ( W_{xi\downarrow}^N x_{t}  + W_{hi\downarrow}^N  h_{t-1} + W_{ci\downarrow }^N  c_{t-1}   + b_{i\downarrow}^N ) \\
        f_{t} &= \sigma ( W_{xf\downarrow}^N x_{t}  + W_{hf\downarrow}^N  h_{t-1} + W_{cf\downarrow}^N  c_{t-1}   + b_{f\downarrow}^N ) \\
        o_{t} &= \sigma (W_{xo\downarrow}^N x_{t}  + W_{ho\downarrow}^N  h_{t-1} + W_{co\downarrow}^N  c_{t}   + b_{o\downarrow}^N)  \\
    \end{split}
\end{equation}
Here $N \in \{L, R\}$ and $\mathbf{U}_N=\{W_{xg\downarrow}^N, W_{hg\downarrow}^N, \\  b_{g\downarrow} ^ N, W_{xi\downarrow}^N, W_{hi\downarrow}^N, 
W_{ci\downarrow }^N, b_{i\downarrow}^N,   W_{xf\downarrow}^N, W_{hf\downarrow}^N,  W_{cf\downarrow}^N,
b_{f\downarrow}^N, \\W_{xo\downarrow}^N, W_{ho\downarrow}^N,  W_{co\downarrow}^N, b_{o\downarrow}^N\}$. 
$\mathbf{U}_L$ and $\mathbf{U}_R$ are model parameters in the top-down Tree LSTM.  

One final note is that the top-down Tree LSTM is enabled by the head propagation mechanism, which allows a head lexicon node to be made available for the root 
constituent node. 
Without such information, it would be difficult to build  top-down LSTM for constituent trees. 

\section{Classification}
We apply the bidirectional Tree LSTM to classification tasks, where the input is a sentence with its binarized constituent tree, and the output is a discrete label.
%$L \in \{0,1,2,3,4\}$. 
Denoting the bottom-up hidden state vector of the root as $\tilde{h}_{ROOT\uparrow}$,  the top-down hidden state vector of the root as $\tilde{h}_{ROOT\downarrow}$ and the top-down hidden state vectors of the input words $x_1$, $x_2$, \dots, $x_n$ as $\tilde{h}_1^{\prime}$, 
$\tilde{h}_2^{\prime}$, \dots, $\tilde{h}_n^{\prime}$, respectively,  we take the concatenation of  $\tilde{h}_{ROOT\uparrow}$, $\tilde{h}_{ROOT\downarrow}$ and the average of $\tilde{h}_1^{\prime}$, 
$\tilde{h}_2^{\prime}$, \dots, $\tilde{h}_n^{\prime}$ as the final representation $h$ of the sentence:
\begin{equation}
 h = \tilde{h}_{ROOT\uparrow} \oplus \tilde{h}_{ROOT\downarrow} \oplus \frac{1}{n} \sum_{i=1}^{n} \tilde{h}_i^{\prime}
\end{equation}
A softmax classifier is used to predict the probability $p_j$ of sentiment label $j$ from $h$ by
\begin{equation}
\label{eq:prob}
\begin{split}
    h_l &= \text{ReLU}( W_{hl} h + b_{hl} )  \\
    P &= \text{\emph{softmax}}( W_{lp} h_l + b_{lp} ) \\
    p_j &= P[j], \\
\end{split}
\end{equation}
where $W_{hl}$, $b_{hl}$, $W_{lp}$ and $b_{lp}$ are model parameters, and ReLU is the rectifier function $f(x) = \max(0, x)$.  During prediction, the largest probability component of $P$ will be taken as the answer.

\section{Training}
We train our classifier to maximize the conditional log-likelihood of %sentiment labels of constituent sentiment trees.  
gold labels of training samples. 
Formally, given a training set of size $|D|$, the training objective is defined by 
\begin{equation}
  L(\theta ) = -\sum_{i=1} ^{ |D|}  \log p_{y_i} + \frac{\lambda}{2} ||\theta||^2,  
\end{equation}
where $\theta$ is the set of model parameters, $\lambda$ is a regularization parameter, 
 $y_i$ is the gold label of the $i$-th training sample and $p_{y_i}$ is obtained according to Equation \ref{eq:prob}.
 For sequential LSTM models,  we collect errors over each sequence.  
 For Tree LSTMs,  we  sum up errors at every node. 

%Our implementation is based on the \emph{CNN} toolkits\footnote[1]{https://github.com/clab/cnn}.   
%In the forwarding phrase, a computation graph is built as a abstract expression tree. 
%During backpropagation,  the  gradients from the top node in the computation graph are propagated down to every bottom node automatically via backpropagation over structures  \cite{goller1996learning}.   
%A node in the computation graph can have multiple parent nodes, and gradients generated from all parent nodes are accumulated together and  passed to child nodes. 

The model parameters are optimized using ADAM \cite{adam} without gradient clipping, with the default hyper-parameters of the AdamTrainer in the 
\emph{Dynet} toolkits.\footnote{https://github.com/clab/dynet}
%stochastic gradient descent with momentum.  The momentum is update by:
%\begin{equation}
%\begin{split}
%   v &= \gamma v + \eta g_\theta  \\
%   \theta &= \theta - v \\
%\end{split}
%\end{equation}
%where $v$ is the current velocity vector, $\eta$ is the learning rate and $\gamma$ is a hyper-parameter for controlling how many history gradients will be incorporated into the current update. 
%We denote the initial learning rate as $\eta_0$, as the number of epochs completed increases, the learning rate decays  in the same way as \newcite{dyer-EtAl:2015:ACL-IJCNLP}. 
We also use dropout \cite{srivastava2014dropout} at lexical input embeddings with a fixed probability $p_{drop}$  to avoid overfitting. $p_{drop}$ is set to 0.5 for all tasks. 

Following \newcite{tai15}, \newcite{li2015tree}, \newcite{zhu15} and \newcite{le15}, we use \emph{Glove-300d}  word embeddings\footnote{http://nlp.stanford.edu/data/glove.840B.300d.zip} to train our model. 
The pretrained word embeddings are fine-tuned for all tasks. 
 Unknown words are handled in two steps. First, if a word is not contained in the pretrained word embeddings,  but its lowercased form exists in the embedding table,   
 we use the lowercase as a replacement. Second,  if both the original word and its lowercased form cannot be found, we treat the word as \emph{unk}.  The embedding vector of the UNK token is initialized as the average of all embedding vectors.

%\begin{table}
%\vspace{0.3cm}
%\begin{center}
%\vspace{0.3cm}
% \begin{tabular}{l}
%  \hline
%  hyper-parameters\\
%  $\lambda = 1e-6$, $\eta_0 = 0.01$, $\gamma = 0.9$ \\
%  $p_{drop} = 0.5$, $p_{unk} = 0.5$\\
%  \hline
% \end{tabular}
% \caption{Hyper-parameter values for  model training. \\}
%\end{center}
%\label{tab:hyper}
%\end{table}
%Table \ref{tab:hyper} shows the hyper-parameters during our model training. 

We use one hidden layer and same  dimensionality settings for both sequential and Tree LSTMs.  
LSTM hidden states are of size 150. 
The output hidden size is 128 and 64 for the sentiment classification task and the question type classification task, respectively.
Each model is trained for 30 iterations. 
%30 iterations are used for training each model.  
The same training procedure is repeated for 5 times using different random seeds, with parameters being evaluated at the end of every iteration on the development set. 
We use the label accuracy to select the best model over the 5 development runs and across the 30 iterations, which is taken for the final test. 

%Our  models were trained  on a single thread on an E7- 4870 2.40GHz CPU. Using the settings discussed in the above, it took one day to finish all 30 iterations for 10 simultaneous experiments.

\section{Experiments}
\subsection{Data}
The effectiveness of our model is tested on a sentiment classification task and a question type classification task. 
\subsubsection{Sentiment Classification}
For sentiment classification, we use the same data settings as \newcite{zhu15}.  
Specifically,  the Stanford Sentiment Treebank \cite{Socher2013EMNLP} is used to train our classification model, which contains sentiments for movie reviews. 
Each sentence is annotated with a constituent tree.%, as shown in Figure \ref{fig:sentitree}.  
Every internal node corresponds to a phrase. 
Each node is manually assigned an integer sentiment label from 0 to 4, which respectively correspond to five sentiment classes: very negative, negative, neutral, positive and very positive.  
The root label represents the sentiment label of the whole sentence. 
%Though, it is also possible to perform binary sentiment classification task on this treebank after label conversions, we focus on the fine-grained sentiment classification problem. 

We  perform both binary classification and fine-grained classification. 
%\begin{figure}[!t]
%    \begin{center}
%        \includegraphics[scale=0.3]{sentitree.pdf}
%        \caption{{ Example sentiment tree. }}
%        \label{fig:sentitree}
%    \end{center}
%    \vspace{-2em}
%\end{figure}
%Table \ref{tab:stat} shows the numbers of root labels and  phrase labels of training, development and test sets,  respectively. 
%The phrase labels include labels from sentences, phrases and words.  
Following previous work, we use labels of all phrases for training. 
 Gold-standard tree structures are used for training and testing \cite{le15,li2015tree,zhu15,tai15}. 
Accuracies are evaluated for both  the sentence root labels and phrase labels.
%at the testing phrase.  
%The accuracy is given by 
%$$acc = \frac{\#correct}{\#total}$$

%\begin{table}[!t]
%	\begin{center}

%	\vspace{0.3cm}
%	\setlength{\tabcolsep}{1.5pt}
%\begin{tabular}{ccccc}
%	\hline
%	Data Set& \#Root5 & \#Phrase5 & \#Root2 & \#Phrase2\\ 
%	\hline
%	train & 8,544 & 318,582  & 6,920  & 84,440\\ 
%	dev & 1,001 & 41,447 & 872 & 11,033\\ 
%	test & 2,210 & 82,600 & 1,821 & 22,451\\
%	\hline
%\end{tabular} 
%	\caption{Statistics of the sentiment treebank. Root5 and Phrase5 denote the numbers of sentences and phrases for the 5-class task, respectively. Similarly, Root2 and Phrase2 denote the same metrics for the binary classification task.\\}
%	\label{tab:stat}
%\end{center}
%\vspace{-3em}
%\end{table}

\subsubsection{Question Type Classification}
For the question type classification task, we use the TREC data \cite{litrec}. 
Each training sample in this dataset contains a question sentence and its corresponding question type. We work on the six-way coarse classification task, where the six question types are \emph{ENTY}, \emph{HUM}, \emph{LOC}, \emph{DESC}, \emph{NUM} and \emph{ABBR}, corresponding to  \emph{ENTITY}, \emph{HUMAN}, \emph{LOCATION}, \emph{DESCRIPTION}, \emph{NUMERIC VALUE} and  \emph{ABBREVIATION}, respectively.  For example, the correct type for the sentence ``What year did the Titanic sink?'' is of the NUM type. The training set consists of 5,452 examples and the test set contains 500 examples. Since there is no development set, we follow  \newcite{clstm}, randomly extracting 500 examples from the training set as a development set. Different from the sentiment treebank, there is no annotated tree for each sentence. Instead, we obtain an automatically parsed tree for each sentence using an open-sourced parser off-the-shelf  (ZPar)\footnote{https://github.com/SUTDNLP/ZPar, version 7.5}. Another difference between the TREC data and the sentiment treebank is that there is only one label, at the root node, rather than a label for each phrase.  

\subsection{Baselines}
We consider two models for our baselines. 
The first is a bidirectional LSTM (\textbf{BiLSTM}) \cite{hochreiter1997long,graves2013hybrid}. 
Our bidirectional constituency Tree LSTM (\textbf{BiConTree}) are compared  with the bidirectional sequential LSTM to investigate the effectiveness of the tree structure. 
For the sentiment task, following \newcite{tai15} and \newcite{li2015tree}, we convert the treebank into sequences to allow the bidirectional LSTM model to make use of every phrase span as a training example. 
The second baseline model is  the bottom-up Tree LSTM model of \newcite{zhu15}. 
We make a contrast between this model and our lexicalized bidirectional models to show the effects of adding head lexicalization and  top-down information flow. 

\subsection{Main Results}
Table \ref{tab:main} shows the main results for the sentiment classification task,  where \textbf{RNTN} is the recursive neural tensor model of \newcite{Socher2013EMNLP}, \textbf{ConTree} and \textbf{DepTree} denote constituency Tree LSTMs and dependency Tree LSTMs,  respectively. 
Our re-implementation of sequential bidirectional LSTM and constituent Tree LSTM \cite{zhu15}  gives comparable results to \newcite{zhu15}'s original implementation.
\begin{table}[!t]
\small
\setlength{\tabcolsep}{2pt}
    \begin{center}
    %\vspace{0.3cm}
        \begin{tabular}{lcccc}
            \hline
            \multirow{2}{*}{Model} &  \multicolumn{2}{c}{5-class} & \multicolumn{2}{c}{binary} \\
            %\hline
             & Root   & Phrase  & Root & Phrase \\
            \hline
            \footnotesize{RNTN\cite{Socher2013EMNLP}} &   45.7 &  80.7 & 85.4  & 87.6\\
             \footnotesize{BiLSTM\cite{li2015tree}}   & 49.8 &   83.3  & 86.7 &  -  \\
             \footnotesize{DepTree\cite{tai15}}  & 48.4 & - & 85.7 & - \\
            \footnotesize{ConTree\cite{le15}}  & 49.9  & - & 88.0  & -\\
            \footnotesize{ConTree\cite{zhu15}} & 50.1   & - & - & -\\
            \footnotesize{ConTree\cite{li2015tree}}   & 50.4 &  83.4 & 86.7 & -\\
            \footnotesize{ConTree\cite{tai15}}  & 51.0 & - & 88.0 & -\\
            \hline 
            \footnotesize{BiLSTM (Our implementation)} & 49.9  & 82.7 & 87.6  &  91.8 \\
            \footnotesize{ConTree (Our implementation)}  &  51.2 & 83.0 &  88.5 & 92.5\\
            \hline
            \footnotesize{Top-down ConTree} & 51.0 & 82.9 & 87.8 & 92.1\\
            \footnotesize{ConTree + Lex}  & 52.8   & 83.2 & 89.2    &  92.3 \\
            \footnotesize{BiConTree} &  \textbf{53.5} & \textbf{83.5} &  \textbf{90.3} & \textbf{92.8}\\
            \hline
        \end{tabular}
         \caption{Test set accuracies for sentiment classification tasks. }
         \label{tab:main}
    \end{center}
    \vspace{-1.5em}
\end{table}

%\begin{table}[!t]
%\small
%\setlength{\tabcolsep}{2pt}
%    \begin{center}
    %\vspace{0.3cm}
%        \begin{tabular}{lcc}
%            \hline
%            Model & Root   & Phrase  \\
%            \hline
%            RNTN\cite{Socher2013EMNLP} &  85.4  & 87.6  \\
%            BiLSTM\cite{li2015tree}   & 86.7 &  -  \\
%            BiLSTM\cite{tai15} & 87.5 & -\\
%            DepTree\cite{tai15}  & 85.7 & - \\
%            ConTree\cite{le15}  & 88.0  & -\\
            %ConTree\cite{zhu15} &  -  & - \\
%            ConTree\cite{li2015tree}   & 86.7 & -  \\
            %ConTree\cite{tai15}  & 88.0 & -\\
           %\hline 
        %    BiLSTM (Our implementation) & 87.6  &  91.8 \\
         %   ConTree  (Our implementation)  &  88.5 & 92.5   \\
         %   \hline 
         %   Top-down ConTree   & 87.8 & 92.1\\
          %  ConTree + Lex  & 89.2    &  92.3\\
          %  BiConTree &  \textbf{90.3} & \textbf{92.8} \\
         %   \hline
        %\end{tabular}
         %\caption{Test set accuracies for the binary sentiment classification task. }
         %\label{tab:mainb}
    %\end{center}
    %\vspace{-2em}
%\end{table}

After incorporating head lexicalization into our constituent Tree LSTM, the fine-grained sentiment classification accuracy increases from $51.2$ to $52.8$, and the binary sentiment classification accuracy increases from $88.5$ to $89.2$, which demonstrates the effectiveness of the head lexicalization mechanism.   

Table \ref{tab:main} also shows that a vanilla top-down ConTree LSTM enabled by head-lexicalization (i.e. the top-down half of the final bidirectional model) alone obtains comparable accuracies to the bottom-up ConTree LSTM model. 
%The main reason is that there is relatively little lexical information at the root level. 
%In addition, 
The \textbf{BiConTree} model can further improve the classification accuracies by 0.7 points (fine-grained) and 1.3 points (binary) compared to the unidirectional bottom-up lexicalized ConTree LSTM model, respectively. 
  
Table \ref{tab:main} includes  5 class accuracies for all nodes. There is no significant difference between different models, as consistent with the observation of \newcite{li2015tree}.  To our knowledge, these are the best reported results for this sentiment classification task until now.
%We are also surprised by the results. There is not a very obvious reason. Maybe both the bidirectional LSTM models and tree LSTM models already reach the top accuracy for this dataset, where there are much more training examples for the phrase level task for reaching a bottleneck away. 

%

%We also evaluate sentiment classification accuracy at the phrase level. However, unfortunately, there is no significant improvement. The accuracy at the phrase level is around 83.0.
%This observation is consistent with the findings reported in \newcite{li2015tree} that Tree LSTMs help identify the sentiment label at the root level, 
%while it could not yield significant improvement at the phrase level. 
Table \ref{tab:trec} shows the question type classification results.
Our final model gives  better results compared to the BiLSTM model and the bottom-up ConTree model, achieving comparable results to the state-of-the-art, which is a SVM classifier with carefully designed features. 
%significantly

\begin{table}[!t]
    \begin{center}
        %\vspace{0.3cm}
        \begin{tabular}{lc}
           \hline
Model 			      	&  Accuracy  \\	
\hline
 Baseline BiLSTM 		      	&  93.8 		\\
 Baseline BottomUp ConTree LSTM  	&  93.4 		\\
 SVM \cite{silva2011symbolic} &  \textbf{95.0} 		\\
 \hline
 Bidirectional ConTree LSTM             	&  94.8		\\
 \hline
\end{tabular}
   \caption{TREC question type classification results.  }
    \label{tab:trec}
    \end{center}
\vspace{-1em}
\end{table}

\begin{table}[!t]
%\small
%\setlength{\tabcolsep}{2pt}
 \begin{center}
 %\vspace{0.3cm}
 \begin{tabular}{lccccc}
  \hline
  %\fontnotesize{Method} & \footnotesize{L} & \footnotesize{R} & \footnotesize{A} & \footnotesize{G}\\
   {Method} & {L} & {R} & {A} & {G}\\
  \hline
 {Root Accuracy} (\%) & 	51.1 & 51.6 & 51.8 &  53.5 \\
  %LEFT &	51.1 \\
  %RIGHT & 51.6 \\
  %AVERAGE & 51.8 \\
  %GATEDLEX  &  53.5 \\
  \hline
 \end{tabular}
\caption{Test set accuracies of four head lexicalization methods on fine-grained classification.}
\label{tab:headlex}
\end{center}
\vspace{-1em}
\end{table}

\subsection{Training Time and Model Size}

Introducing head lexicalization and bidirectional extension to the model increases the model complexity. In this section, we analyze our model in terms of training time and model size on the fine-grained sentiment classification task. 

\begin{table}[!t]
 \small
 \begin{center}
 %\vspace{0.3cm}
 \begin{tabular}{llll}
  \hline
  Model &  ConTree & ConTree+Lex & BiConTree \\
  \hline
  Time (s) & 4,664 & 7,157 & 11,434 \\
  \hline
 \end{tabular}
\caption{Averaged training time over 30 iterations.\\}
\label{tab:time}
\end{center}
%\vspace{-3em}
\end{table}
We run all the models using an i7-4790 3.60GHz CPU with a  single thread. 
Table \ref{tab:time} shows the average running time for different models over 30 iterations. The baseline \textbf{ConTree} model takes about 1.3 hours to finish the training procedure. \textbf{ConTree+Lex} takes about 1.5 times longer than \textbf{ConTree}. \textbf{BiConTree} takes  about 3.2 hours, which is about 2.5 times longer than that of  \textbf{ConTree}. 
%On small dataset, it seems that the training time for \textbf{BiConTree} model is still acceptable. For example, it can process one training example on the SST dataset using only 0.04 seconds.  For large dataset, the \textbf{ConTree+Lex} model is a good compromise for performance and speed. 

\begin{table}[!t]
\small
 \begin{center}
 %\vspace{0.3cm}
  %\setlength{\tabcolsep}{2.5pt}
 \begin{tabular}{llrc}
  \hline
  Model &  $|h|$ &  \#Parameter & Accuracy (\%) \\
  \hline
  ConTree & 150 & 538,223 & 51.2 \\
  \hline
  ConTree+Lex &	75	& 376,673 & 51.5 \\
ConTree+Lex&	150	& 763,523 & 52.8  \\
ConTree+Lex&	215	& 1,253,493 & 52.5  \\
ConTree+Lex &	300	& 2,110,973 & 51.8 \\
\hline 
BiConTree  &	75	& 564,923  & 52.6  \\
BiConTree  &	150	& 1,297,523  & 53.5 \\
  \hline
 \end{tabular}
\caption{Effect of model size.\\}
\label{tab:modelsize}
\end{center}
\vspace{-1.5em}
\end{table}

Table \ref{tab:modelsize} compares the model sizes. We did not count the number of parameters in the lookup table since these parameters are the same for all models. Since the size of LSTM models mainly depends on the dimensionality of the state vector $h$, we change the size of $h$ for comparing the effect of model size. When $|h|=150$, the model size of the baseline model \textbf{ConTree} is the smallest, which consists of about 538K parameters. The model size of \textbf{ConTree+Lex} is about 1.4 times as large as that of the baseline model. The bidirectional model \textbf{BiConTree} is the largest, which is about 1.7 times as large as that of the \textbf{ConTree+Lex} model.  However, this parameter set is not very large compared to the modern memory capacity, even for a computer with 16GB RAM. 
In conclusion, in terms of both time, model size and accuracy, head lexicalization method is a good choice.

Table \ref{tab:modelsize} also helps to clarify whether the gain of the \textbf{BiConTree} model over the \textbf{ConTree+Lex} model is from the top-down information flow or more parameters. 
For the same model, increasing the model size can improve the performance to some extent. For example, doubling the size of $|h|$ ($75\rightarrow150$) increases the performance from $51.5$ to $52.8$ for the \textbf{ConTree+Lex} model. Similarly, we boost the performance of the \textbf{BiConTree} model when doubling the size of $|h|$ from $75$ to $150$.  However, continuing doubling the size of $|h|$ from $150$ to $300$ empirically decreases the performance of the \textbf{ConTree+Lex} model. %This may due to overfit the dataset with a relatively simple model but with large parameters.   
%\\
%However, better model are easier to tune and get a better performance than a carefully fine-tuned worse model. 
%For different models, 
The \textbf{BiConTree} model with $|h|=75$ is much smaller than the \textbf{ConTree+Lex} model with $|h|=150$ in terms of model size. However the performance of these two models is quite close, which indicates that top-down information is useful even for a small model. We also run a \textbf{ConTree+Lex} model with $|h|=215$, the model size of which is similar to that of the \textbf{BiConTree} model with $|h|=150$. The performance of the \textbf{ConTree+Lex} model is still worse than the \textbf{BiConTree} model ($52.5$ v.s. $53.5$), which shows the effectiveness of top-down information. %In addition, more advanced models are easier for  obtaining better performance than a carefully fine-tuned baseline model. 

\subsection{Head Lexicalization Methods}
In this experiment, we investigate the effect of our head lexicalization method over heuristic baselines. 
We consider three baseline methods, namely \emph{L}, \emph{R} and \emph{A}.  
For \emph{L},  a parent node accepts lexical information of its left child while ignoring the 
right child.  Correspondingly,  for \emph{R},  a parent node accepts lexical information of its right child while ignoring the 
left child.   For \emph{A},  a parent node averages the lexical vectors of the left and right child.

Table \ref{tab:headlex} shows the accuracies on the test set,  where \emph{G} denotes our method described in Section \ref{sec:headlex}. 
\emph{R} gives better results compared to \emph{L} due to relatively more  right-branching structures in this treebank.%, as shown in Figure \ref{fig:sentitree}.
%Methods which considers both left child and right child perform better than those depending on a single branching.  
A simple average yields similar results compared with right branching. 
In contrast, \emph{G} outperforms \emph{A} method by considering the relative weights of each branch according to tree-level contexts. 

\begin{figure}[!t]
%\centering
\begin{subfigure}[b]{0.24\linewidth}
    %\begin{center}
    \includegraphics[scale=0.32]{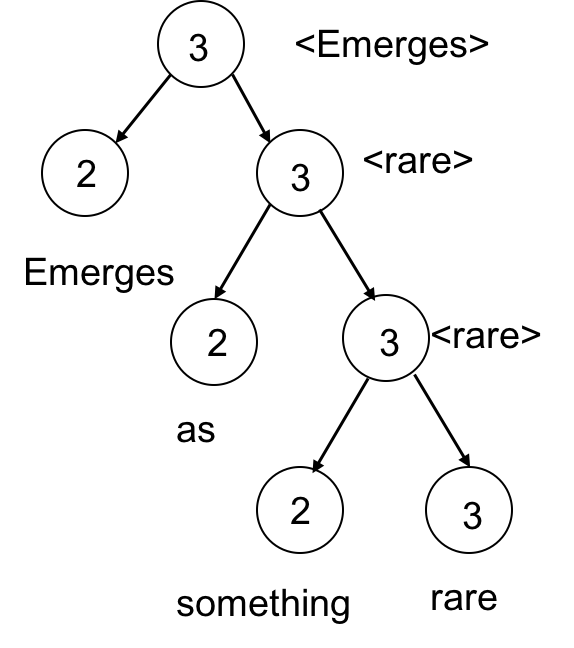}
        \caption{``Emerges as something rare''.}
        \label{fig:headlexeg1}
    %\end{center}
\end{subfigure}
%\par\bigskip
\hspace{0.5in}
%\begin{subfigure}[b]{0.22\linewidth}
\begin{subfigure}[b]{0.22\textwidth}
    
        \includegraphics[scale=0.32]{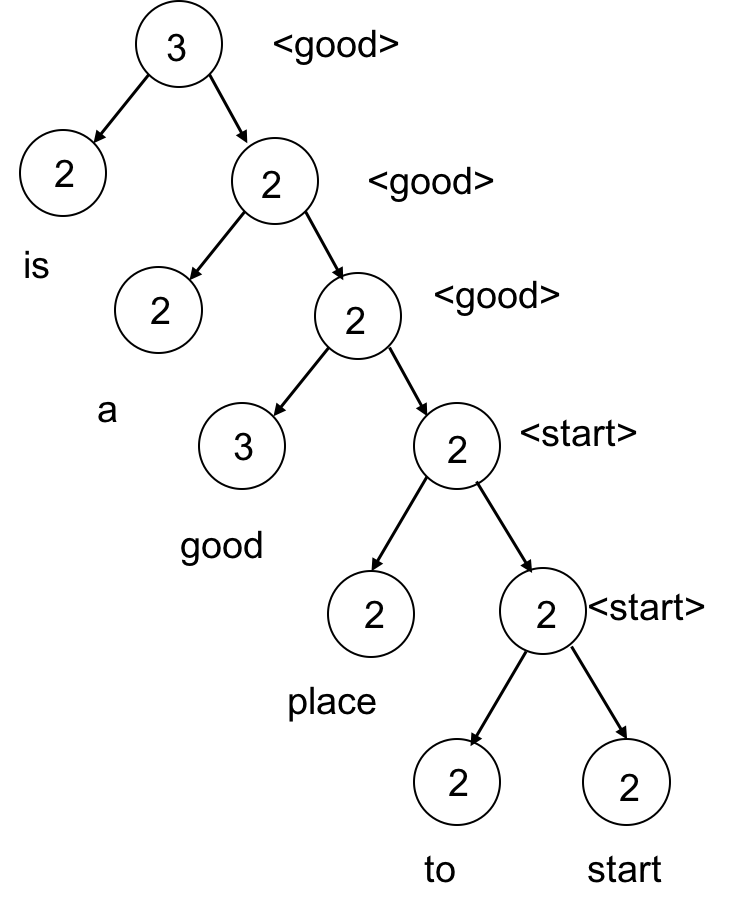}
        \caption{``is a good place to start''.}
        \label{fig:headlexeg2}
\end{subfigure}
\par\bigskip
\begin{subfigure}[b]{0.45\textwidth}
    %\begin{center}
        \includegraphics[scale=0.32]{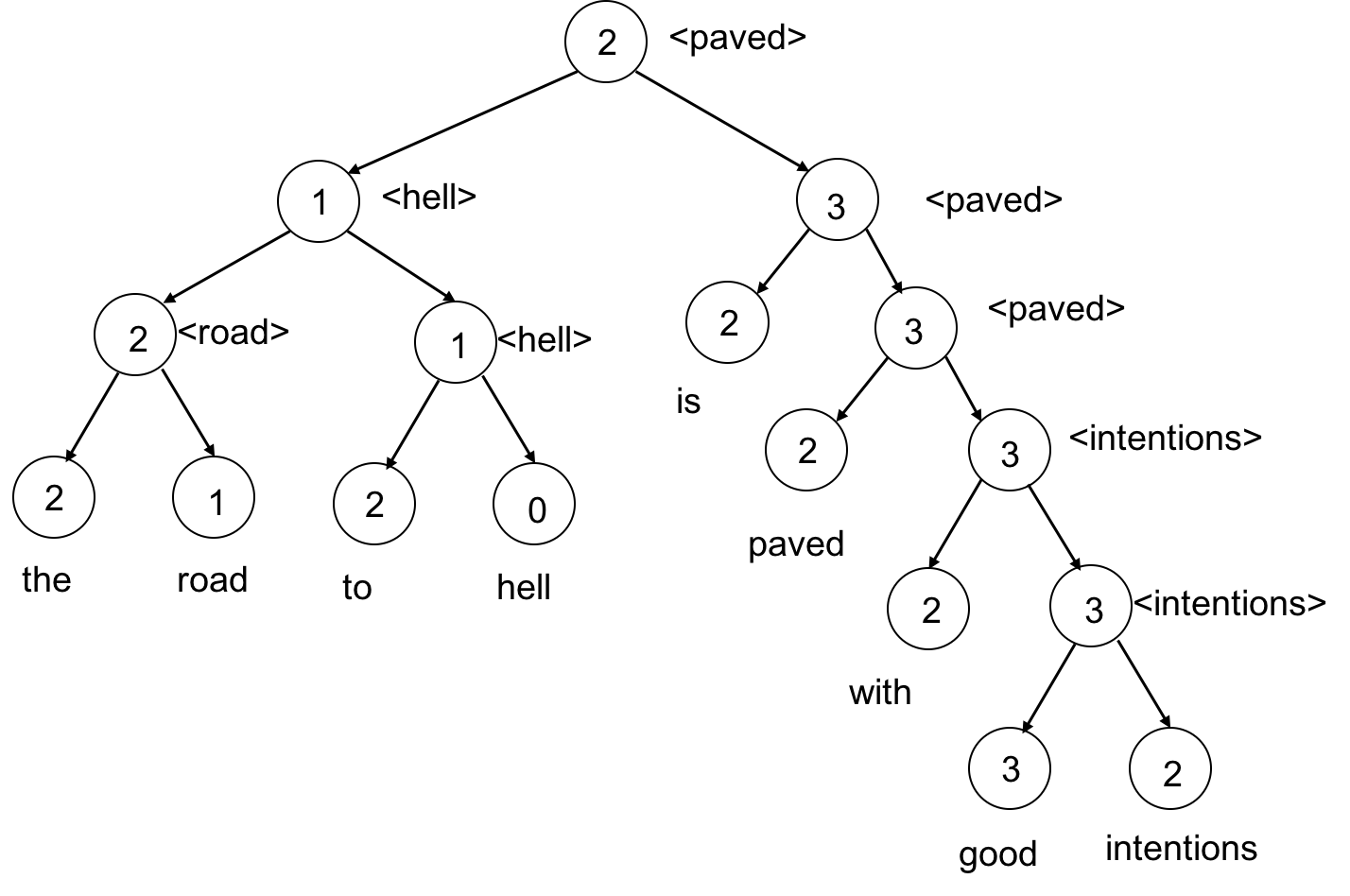}
        \caption{``the road to hell is paved with good intentions''.}
        \label{fig:headlexeg3}
    %\end{center}
\end{subfigure}
\caption{Visualizing head words found automatically by our model.}
\label{fig:vis}
\end{figure}

We then investigate what lexical heads can be  learned by \emph{G}. Interestingly, the lexical heads contain both syntactic and sentiment information: while some heads correspond well to Collins syntactic rules \cite{collins2003head}, others are driven by subjective words.
Compared to Collins' rules, our method found 30.68\% and 25.72\% overlapping heads on the dev set and test set, respectively. 

Figure \ref{fig:vis} shows some visualization based on the cosine similarity between the head lexical vector and its children. We decide the head of a node by choosing  the head of the child that gives the largest similarity value. Some examples are shown below, where $<>$ is used to indicate head words. Sentiment labels (e.g. 2, 3)  are also included. In Figure \ref{fig:headlexeg1}, ``Emerges'' is the syntactic head word of the whole phrase, which is consistent with Collins-style head finding. However, ``rare'' is the head word of the phrase ``something rare'', which is different from the syntactic head, but rather sentiment-related. Similar observations are found in Figure \ref{fig:headlexeg2}, where ``good'' is the head word of the whole phrase, rather than the syntactic head ``place''. The sentiment label of ``good'' and the sentiment label of the whole phrase are both 3.  Figure \ref{fig:headlexeg3} shows more complex interactions between syntax and sentiment for deciding the head word. 

\subsection{Error Analysis}

 Table \ref{tab:example} shows  some example sentences incorrectly predicted by the baseline bottom-up tree model, but correctly labeled by our final model.
 %for qualitative discussions. 
 \begin{table}[!t]
 \small
 \setlength{\tabcolsep}{1.5pt}
 \begin{center}
 \vspace{0.3cm}
 \begin{tabular}{llcc}
  \hline
ID & sentence 					               & baseline & our model \\
\hline
1  & \tabincell{l}{{\color{red}{Gloriously}} goofy {\color{blue}{(}} and \\gory {\color{red}{)}} midnight movie stuff .} & negative& positive \\
\hline
2  & \tabincell{l}{The film is really {\color{blue}{not}} \\so much {\color{blue}{bad}} as {\color{blue}{bland}} .} & positive & negative\\
\hline
%3  & \tabincell{l}{The film 's {\color{red}{real}} {\color{red}{appeal}} \\wo {\color{blue}{n't}} be to Clooney fans \\or adventure buffs , but to \\moviegoers who enjoy thinking \\about compelling questions \\with {\color{blue}{no}} easy answers . } & negative & positive \\
%\hline
%3  & \tabincell{l}{This film is so {\color{red}{slick}} , \\ superficial and trend-hoppy , \\that it 's easy to imagine that\\ a {\color{red}{new}} software program \\{\color{blue}{spit}} {\color{blue}{out}} the screenplay .} & positive & negative \\
%\hline
3 & \tabincell{l}{This is a {\color{red}{good}} movie \\in spurts , but when \\it does n't work , \\it 's at important times .} & positive & neutral \\
\hline
%5  & \tabincell{l}{The reason this picture works\\ better than its predecessors\\ is that Myers is no longer\\ simply spoofing the \\mini-mod-madness\\ of '60s spy movies .} & negative & positive\\
%  \hline
 \end{tabular}
\caption{Example output sentences.}
\label{tab:example}
\end{center}
\vspace{-2em}
\end{table}
%\subsection{Analysis}
The head word of sentence \#1  by our model is ``Gloriously'', which is consistent with the sentiment of the whole sentence.  This shows how head lexicalization can affect sentiment classification results. Sentences \#2 and \#3 show the usefulness of top-down information for complex semantic structures, where compositionality exhibits subtle effects. With top-down LSTM, the models can predict the sentiment labels correctly.  Our final model improves the results  for the `very negative' and `very positive' classes by 10\% and 11\%, respectively. It also boosts the accuracies for sentences with negation (e.g. ``not'', ``no'', and ``none'') by 4.4\%.

\begin{figure}[!t]
    \begin{center}
        \includegraphics[scale=0.62]{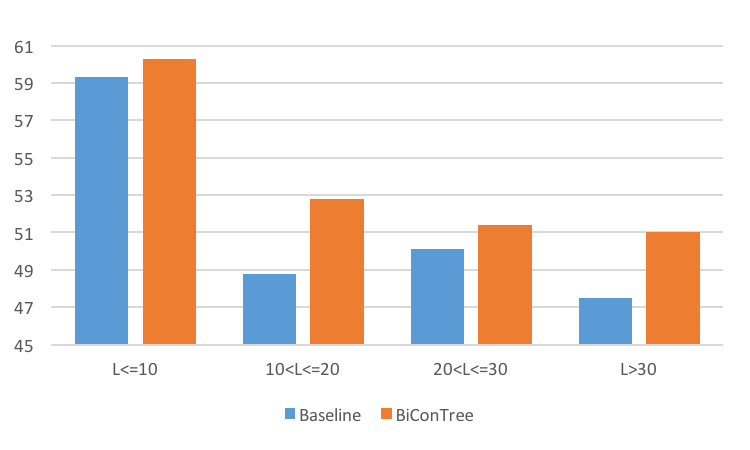}
        \caption{ Distribution of 5-class accuracies at the root level according to the sentence length.}
        \label{fig:lendist}
    \end{center}
    %\vspace{-2em}
\end{figure}

Figure \ref{fig:lendist}  shows the accuracy distribution according to the sentence length.  We find that our model can improve the classification accuracy for longer sentences ($>$30 words) by 3.5 points compared to the baseline ConTree LSTM of \newcite{zhu15}, which demonstrates the strength of our model for handling long range information.  %For example,  for sentence \#3, the baseline model recognizes that there is a positive word ``funny'', and thereby labels the sentence as positive.  
By considering bidirectional information over tree structures, our model is aware of more complicate contexts for making better predictions.  

%We add three example sentences and make sentiment words colorful. We also add a classification confusion matrix to show our final model can improve the results by 10\% and 11\% for the 'very negative' and 'very positive' classes compared to the baseline model. We also find our method can improve the classification accuracy of  sentences which contains negative words including ``not'', ``no'', ``n't'' and ``none'' by 4.4\%.

\subsection{Parser Reranking}

 \begin{table}[!t]
\small
 \begin{center}
 %\vspace{0.3cm}
 \begin{tabular}{l|l}
  \hline
   Model & $F_1$ \\
   \hline 
   %CVG (ReNN) (\newcite{socher-EtAl:2013:ACL2013})  & 85.0 \\
   %CVG (SU-RNN) (\newcite{socher-EtAl:2013:ACL2013}) & 90.4 \\
   %\hline
   %LSTM+A (\newcite{vinyals2015grammar})  & 88.3 \\
   %LSTM+A (\newcite{vinyals2015grammar}) Ensemble & 90.5 \\
   %LSTM+A +HC (\newcite{vinyals2015grammar}) & 92.1 \\
   %Discriminative (\newcite{dyer2016recurrent})  & 91.7 \\
   %Generative (\newcite{dyer2016recurrent}) Rerank & 93.3 \\
   %LSTM-LM (\newcite{charniakparsing})  & 92.6 \\
   %LSTM-LM +HC (\newcite{charniakparsing}) Ensemble & 93.8 \\
  %\hline
   Baseline (\newcite{zhu2013fast}) & 90.40 \\
   ConTree  & 90.70 \\
   ConTree+Lex & 90.83 \\
   \hline
   Our 8-best Oracle & 92.59 \\
   
  \hline
 \end{tabular}
\caption{Neural reranking results on WSJ test set.}
\label{tab:rerankres}
\end{center}
%\vspace{-3em}
\end{table}
Our main results are obtained on semantic-driven tasks, where the automatically-learned head words contain mixed syntactic and semantic information. 
To further investigate the effectiveness of automatically learned head information on a pure syntactic task, we additionally conduct a simple parser reranking experiment. 
The standard PTB \cite{marcus1993building} split \cite{collins2003head}  are used, and the same off-the-shelf ZPar model is adopted for our baseline. A standard interpolated reranker \cite{zhu-EtAl:2015:ACL-IJCNLP,zhou-EtAl:2016:P16-12} is adopted for scoring 8-best trees. For each tree y of sentence x, we follow \newcite{socher-EtAl:2013:ACL2013} to define $f(x, y; \Theta)$ as the sum of scores of each constituent node, 
\begin{equation}
f(x, y; \Theta) = \sum \limits_{r \in \text{node}{(x, y)}} {Score(r; \Theta)} 
\end{equation}
%where the tree-LSTM score is derived from all nodes: 
Without loss of generality, we take a binary node as an example. Given a node A, suppose that its two children are B and C. Let the learned composition state vectors of A, B and C by our proposed Tree-LSTM model be  $N_A$, $N_B$ and $N_C$, respectively. The head word vector of node A is ${H}_A$. $Score(A; \Theta)$ is defined as:
\begin{equation}
\small
    \begin{split}
          &O_{(A\rightarrow BC)} = ReLU(W_{s}^{L}N_B + W_{s}^{R}N_C + W_{s}^{H} H_A + b^s) \\
    &\text{Score}_{(A\rightarrow BC)}  = \log(softmax(O_{(A\rightarrow B\ C)}))[A],
    \end{split}
\end{equation}
where $W_{s}^{L}$, $W_{s}^{R}$ and $b^s$ are model parameters, which are trained using a margin loss.   

Table \ref{tab:rerankres} shows the reranking results on WSJ test set. 
The baseline F1 score is 90.40. Our \textbf{ConTree} improves   the baseline model from $90.40$ to $90.70$. 
Using \textbf{ConTree+Lex} model can further improve the performance ($90.70\rightarrow 90.83$). This serves as an evidence that automatic heads can also be useful to some extent for a syntactic task, despite that our main motivation is more semantic-driven.  
%In total, our final reranker outperform the baseline model by 0.43 F1 score, which show the potential of Tree-LSTMs based reranker. %We also compare our results with other neural reranking parsers (the first block). Our reranker is comparable to the model of \newcite{socher-EtAl:2013:ACL2013}, which is based on recursive composition vectors. 
%Our model is about the same as 
Among neural rerankers, out model outperforms 
\newcite{socher-EtAl:2013:ACL2013}, but is not as good as current state-of-the-art models, including sequence-to-sequence based LSTM language models \cite{vinyals2015grammar,charniakparsing} and recurrent neural network grammars \cite{dyer2016recurrent}, due to a low baseline oracle and simple reranking models\footnote{\newcite{dyer2016recurrent} employs 2-layerd LSTMs with input and hidden dimensions of size 256 and 128, respectively,  and \newcite{charniakparsing} use 3-layered LSTMs with both the input and hidden dimensions of size 1500.}. Nevertheless, it serves our goal of contrasting the tree LSTM models. 

%We think the reasons are two folds. 
%We work on a smaller n-best list. In our experiments, we set $n=8$ while \newcite{charniakparsing} use setting with $n=50$, \newcite{dyer2016recurrent} search over 100 samples. 

%Second, we use a small neural network due to the limit of available hardware computing resources. \newcite{dyer2016recurrent} employs 2-layerd LSTMs with input and hidden dimensions of size 256 and 128, respectively,  and \newcite{charniakparsing} use 3-layered LSTMs with both the input and hidden dimensions of size 1500.  Whether large Tree-LSTMs model is better  than LSTMs remains an interesting problem to explore. 

\section{Conclusion}
We showed that existing LSTM models for constituent tree structures  are limited by not considering direct lexical input in the computing of cell values for non-leaf constituents, and proposed 
a head-lexicalization method to address this issue. 
Learning the heads of constituent automatically using a neural model, our lexicalized tree LSTM is applicable to arbitrary binary  branching  
trees in form of CFG, and is formalism-independent.  In addition, lexical information on the root further allows a top-down  extension to the model, resulting in a bi-directional constituent Tree LSTM. 
Experiments on two well-known datasets show that head-lexicalization improves the unidirectional Tree LSTM model, and bidirectional Tree LSTM gives superior labeling results compared with both unidirectional
Tree LSTMs and bidirectional sequential LSTMs. We release our code under GPL at XXX. 

\bibliography{acl2012}
\bibliographystyle{acl2012}

\appendix 

\end{document}